\documentclass[runningheads]{llncs}

\usepackage{eccv}

\usepackage{eccvabbrv}

\usepackage{graphicx}
\usepackage{booktabs}
\usepackage{threeparttable}
\usepackage{multirow}
\usepackage{arydshln}
\usepackage{xcolor}
\usepackage[table]{xcolor}
\usepackage{pifont}
\usepackage{bbding}
\usepackage{makecell}
\usepackage{amsmath}
\usepackage{amssymb}

\usepackage[accsupp]{axessibility}  

\usepackage{orcidlink}

\begin{document}

\title{PriorMaskMap: Robust Online Vectorized Map Construction with Biased Priors}

\titlerunning{PriorMaskMap}

\author{ Haoming Xu\inst{1,2} \and Wei Li\inst{1,3} \textsuperscript{*} \and Yu Hu\inst{1,3} } \authorrunning{H. Xu et al.} \institute{ The Research Center for Intelligent Computing Systems, SKLP, Institute of Computing Technology, Chinese Academy of Sciences, Beijing 100190, China \and Hangzhou Institute for Advanced Study, University of Chinese Academy of Sciences, Hangzhou 310024, China \and University of Chinese Academy of Sciences, Beijing 100049, China\\ \email{xuhaoming24@mails.ucas.ac.cn\\ liwei2019@ict.ac.cn\\ huyu@ict.ac.cn}\\ \textsuperscript{*}Corresponding author}

\renewcommand{\thefootnote}{}
\footnotetext{\textsuperscript{*} Corresponding author.}
\maketitle

\begin{abstract}

Online vectorized map construction provides autonomous vehicles with essential, real-time semantic and geometric scene understanding. While leveraging prior maps presents an opportunity to improve accuracy, they are often biased due to outdated information or temporary road changes. Existing methods, which typically treat priors as fully reliable or only consider limited noise patterns, suffer severe performance degradation when these priors are imperfect. To address this, we propose PriorMaskMap, a robust framework for prior-guided vectorized map construction. Its first innovation is a Prior-Aware Confidence Estimator (PACE) that evaluates the input rasterized map against current visual observations. It assigns confidence scores to regions classified as reliable, erroneous, or blank, creating a foundational mask that guides two subsequent branches. In one branch, the Dual-Prior Query Generator (DPQG) selectively filters and embeds both rasterized and vectorized priors to initialize informative queries from trustworthy map structures. In the other, the Confidence-Guided Prior Fusion (CGPF) module spatially propagates and refines the mask, using the refined version to govern the fusion of BEV features with the prior map, thereby providing rich contextual representations for the map decoder. Evaluation on the nuScenes dataset shows that PriorMaskMap achieves superior mapping precision over existing approaches and exhibits marked robustness when provided with priors containing apparent errors. Our code will be released at https://github.com/healenrens/PriorMaskMap.

\end{abstract}
\section{Introduction}

The online construction of vectorized high-definition (HD) maps is critical for autonomous driving, providing precise lane geometry and topological relationships for downstream tasks such as trajectory planning and decision-making. While current methods have made significant progress in building maps directly from multi-view images \cite{MapTR,liao2025maptrv2,ding2023pivotnet,liu2024leveraging,yang2024mgmapnet}, their performance can degrade noticeably in challenging scenarios such as poor lighting, heavy occlusions, or at long distances. Fortunately, the increasing availability of offline maps allows them to provide beyond-line-of-sight information as a valuable prior, thereby enabling more accurate mapping in those situations where pure visual perception struggles.

Consequently, a series of prior-enhanced online mapping methods have been developed \cite{xiong2023neural,jiang2024p,peng2025uni,Sun_2025_WACV,zeng2024driving}. These approaches vary in terms of the prior map format and its fusion strategy. Among them, P-MapNet \cite{jiang2024p} designs a pipeline that integrates both standard-definition (SD) and HD map as priors. It fuses the SD prior features with sensor-derived features in the bird's-eye-view (BEV) space using an attention mechanism, while HD map priors are incorporated through a masked autoencoder for refining the output. Uni-PrevPredMap \cite{peng2025uni} employs HD maps as priors by storing them in a tile-indexed map processor. This design enables the model to selectively leverage the stored prior information, where retrieved map vectors are rasterized and fused with BEV features while simultaneously serving to initialize the decoder's queries.

However, prior maps can contain significant biases and errors due to outdated information, large-scale road renovations, or sensor misalignment. Such imperfections can severely degrade the performance of prior-enhanced methods that fully trust the map input \cite{11097394}. Recent works like MapEX \cite{Sun_2025_WACV} and Uni-PrevPredMap \cite{peng2025uni} have recognized and attempted to mitigate this problem. MapEX \cite{Sun_2025_WACV} encodes the prior map into queries and interacts them with BEV features via cross-attention, serving as input to the map decoder. Uni-PrevPredMap \cite{peng2025uni} introduces a tri-mode operational optimization paradigm to reduce dependence on map fidelity assumptions. 
While both methods improve robustness to imperfect priors, explicit modeling and propagation of spatially varying prior reliability for confidence-guided feature fusion remain underexplored.
In real-world driving scenarios, discrepancies between offline maps and actual observations are often more complex, involving structural deviations at the lane-level which may affect multiple adjacent lanes due to road reconstruction \cite{plachetka2020terminology,lambert2021trust}. In such cases, explicitly identifying reliable and erroneous regions in the prior map becomes even more valuable, but this critical capability has not been adequately explored in existing methods.

\begin{figure}[t] 
  \centering
   \includegraphics[width=\linewidth]{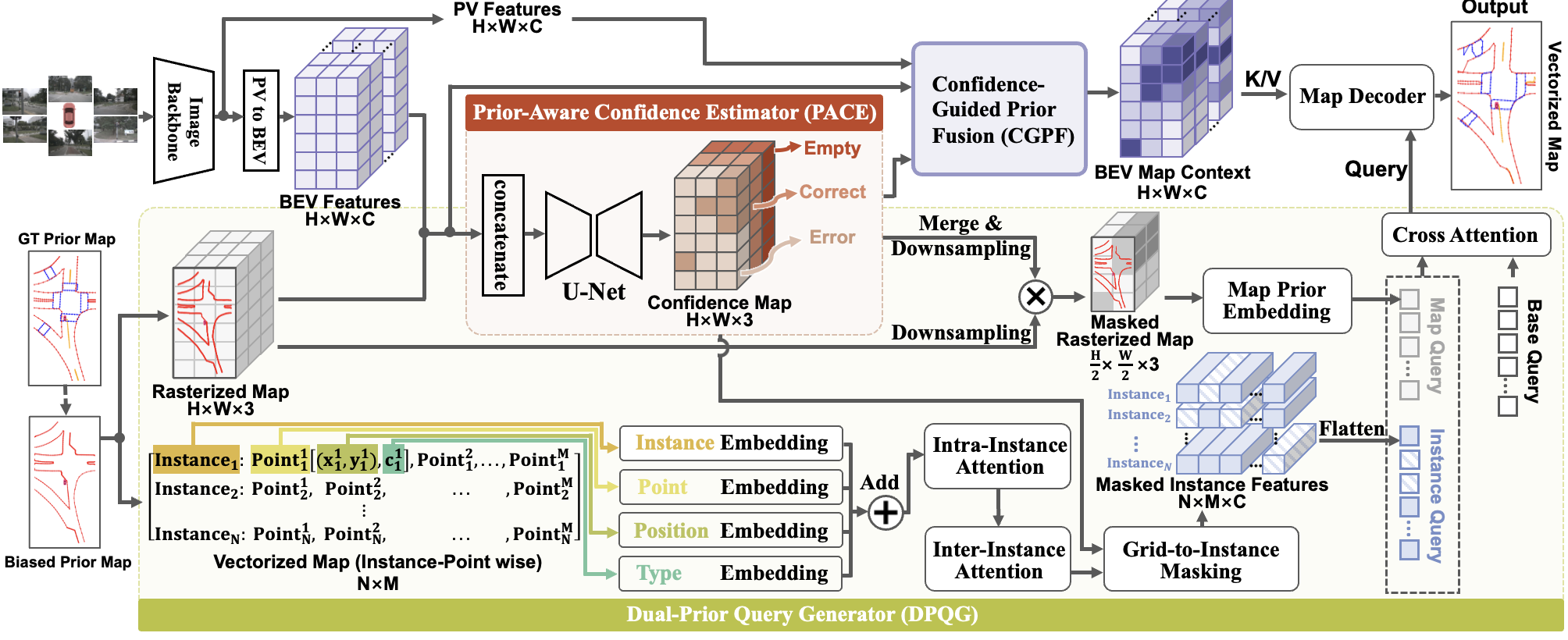}
   \caption{Overall framework of PriorMaskMap. Given multi‑view images and a biased prior map, PACE predicts grid‑wise confidence over the rasterized prior. DPQG converts confident vector and raster structures into prior‑aware queries, while CGPF diffuses and refines the confidence field to produce a BEV map context for the decoder.}
   \label{fig:framework}
\end{figure}

To address this limitation, we propose PriorMaskMap as \cref{fig:framework}, a robust framework for online vectorized map construction that selectively leverages trustworthy regions within prior maps. We first design a Prior-Aware Confidence Estimator (PACE), which classifies each grid of the rasterized prior map as reliable, erroneous, or blank based on current visual observations, and outputs a spatially-differentiated confidence map. This confidence map then acts as a soft mask in our Dual-Prior Query Generator (DPQG), selectively gating the influence of the prior during query initialization. The DPQG is designed with dedicated encoding branches for rasterized and vectorized prior maps, respectively. The resulting embeddings from both branches are merged and interact through cross-attention with a set of learnable queries to decode desirable map elements. Furthermore, we introduce a Confidence-Guided Prior Fusion (CGPF) module, which refines the initial confidence through spatial propagation and uses it to adaptively fuse the prior map with BEV features. This process produces rich contextual representations that serves as Keys and Values for the transformer-based map decoder, effectively grounding the decoding process in both reliable prior knowledge and current sensor observations.

In summary, our contributions are threefold:
\begin{itemize}
\item We propose PriorMaskMap, a novel framework for robust vectorized map construction, built around a core Prior-Aware Confidence Estimator (PACE) that adaptively identifies trustworthy regions in biased prior maps and generates a corresponding grid-wise confidence map.

\item We introduce two key components guided by the confidence map to enhance both map element queries and feature fusion: a Dual-Prior Query Generator that produces gated query embeddings from both rasterized and vectorized priors, and a Confidence-Guided Prior Fusion module that spatially propagates and refines the initial confidence to guide the fusion of BEV features with the prior.

\item We extend the nuScenes dataset with a supplementary set of biased vectorized maps that capture real-world discrepancies, specifically simulating structural deviations across multiple lanes. Experiments show that PriorMaskMap achieves higher accuracy and better robustness against inaccurate priors compared to existing methods.


\end{itemize}

\section{Related Work}
\subsection{Online HD Map Construction}

Early online HD mapping methods treated map construction as BEV segmentation followed by vectorization. HDMapNet \cite{li2022hdmapnet} first performs BEV semantic segmentation and fits lane geometries, which is effective but computationally expensive. Later works move to query\mbox{-}based transformers that directly predict vectorized maps. VectorMapNet \cite{liu2023vectormapnet} uses a two\mbox{-}stage transformer that detects element boxes and then autoregressively decodes polylines. MapTR \cite{MapTR} adopts a DETR\mbox{-}style design to output each lane or boundary as an ordered point set, and MapTRv2 \cite{liao2025maptrv2} enhances it with 3D geometric reasoning and multi\mbox{-}view cues. PivotNet \cite{ding2023pivotnet} instead learns an ordered list of pivot points per lane to better capture complex shapes and topology.

Building on this line, recent models further boost accuracy and robustness \cite{tang2025priorfusion}. MapNeXt \cite{li2024mapnext} shows that scaling model capacity and training leads to strong gains on vectorized map detection. MGMap \cite{liu2024mgmap} learns BEV masks that highlight informative regions and uses them to guide query decoding, improving fine\mbox{-}grained lane geometry under sparse supervision. Mask2Map \cite{choi2024mask2map} combines segmentation and queries by predicting instance\mbox{-}level BEV masks and then generating vectors from them. 

\subsection{Context\mbox{-}Enhanced Online Map Construction}

A complementary line of work exploits temporal context for stable online mapping. SQD\mbox{-}MapNet \cite{wang2024stream} reuses previous outputs via streaming query denoising, MapTracker \cite{chen2024maptracker} propagates latent queries across frames to fuse past BEV features and vectorized maps, ADMap \cite{hu2024admap} adopts a disturbance\mbox{-}rejection training paradigm to handle imperfect sensors and moving obstacles, and EAN\mbox{-}MapNet \cite{xiong2024ean} designs efficient query layouts to accelerate inference while preserving detail.

Many methods further enhance mapping with priors. P\mbox{-}MapNet \cite{jiang2024p} injects fixed HD/SD map priors into the decoder to improve long\mbox{-}range structure, while PriorMapNet \cite{wang2024priormapnet} and PriorDrive \cite{zeng2024driving} encode vectorized map priors into the query space for lane retrieval and ordering. Temporal priors leverage the vehicle’s own history: HRMapNet \cite{zhang2024enhancing} accumulates past predictions into a lightweight raster to augment decoding and initialize queries, and PrevPredMap \cite{peng2025prevpredmap} and MemFusionMap \cite{song2025memfusionmap} forward previous maps or features to handle occlusions and stabilize updates. External environment priors extend this idea further: PreSight \cite{yuan2024presight} introduces city\mbox{-}scale neural radiance priors for geometric regularization, AerialHDMapper \cite{xie2025aerialhdmappper} uses aerial imagery for long\mbox{-}range and cross\mbox{-}view constraints, and ImagineMap \cite{ji2024imaginemap} and PriorFusion \cite{tang2025priorfusion} enhance structure from SD maps and fuse multiple prior sources.

However, real priors are often biased or outdated. Empirical studies \cite{bateman2024exploring} show that prior\mbox{-}informed models are sensitive to misalignment and map changes, motivating explicit change awareness and alignment. ExelMap \cite{wild2024exelmap} targets explainable element\mbox{-}level change detection, RTMap \cite{du2025rtmap} couples online mapping with recursive change detection for self\mbox{-}correction, and ArgoTweak \cite{wild2025argotweak} proposes structured prior\mbox{-}driven self\mbox{-}updating at the database level. These systems retain global and long\mbox{-}range benefits from priors while learning to align, down\mbox{-}weight, or override them, and our work follows this direction by focusing on biased prior maps and designing mechanisms that safely exploit their structure while actively correcting their errors.

\section{Methodology}

Given multi\mbox{-}view images and a biased prior map, our goal is to construct an online vectorized HD map that selectively uses the prior when reliable and otherwise relies on current observations. As shown in ~\cref{fig:framework}, PriorMaskMap consists of three main components. The Prior\mbox{-}Aware Confidence Estimator predicts a per\mbox{-}grid confidence map over the rasterized prior. The Dual\mbox{-}Prior Query Generator uses this confidence to select reliable vector and raster structures and converts them into map\mbox{-}aware queries for the decoder. The Confidence\mbox{-}Guided Prior Fusion diffuses and refines the confidence field on the BEV grid and fuses it with visual features to produce a prior\mbox{-}aware BEV context. We describe these components successively.

\subsection{Prior\mbox{-}Aware Confidence Estimator (PACE)}
Recent architectures often assume that HD or SD prior maps are correct \cite{zhang2024enhancing,jiang2024p}, which can mislead downstream fusion under road reconstruction \cite{11097394}. We relax this assumption with a Prior\mbox{-}Aware Confidence Estimator (PACE) that predicts a dense confidence map and measures spatial consistency between the prior and current observations. High\mbox{-}confidence regions of the prior are preserved, while low\mbox{-}confidence regions fall back to evidence from the current view.

To assess map reliability from actual observations we use BEV features as the decision anchor. The BEV feature map $F_{\mathrm{BEV}}\in\mathbb{R}^{H\times W\times C}$ provides a unified top\mbox{-}down representation, where $H$ and $W$ denote the spatial resolution and $C$ is the channel dimension. We rasterize the prior map into $M_{R}\in\mathbb{R}^{H\times W\times 3}$ and concatenate $F_{\mathrm{BEV}}$ and $M_{R}$ along the channel dimension. The concatenated tensor is fed into a U\mbox{-}Net head \cite{ronneberger2015u} that predicts reliability on the raster grid. For each grid $u$, the network outputs a confidence map $M_{\mathrm{conf}}(u)\in\mathbb{R}^{3}$ for the classes reliable, erroneous, and empty, forming a confidence map $M_{\mathrm{conf}}\in\mathbb{R}^{H\times W\times 3}$ that guides subsequent selection and fusion.

To supervise PACE, we generate targets with a patch\mbox{-}level hybrid scheme based on rasterized maps. Let $M_{\mathrm{gt}}$ denote the ground\mbox{-}truth vectorized map and $M_{\mathrm{err}}$ the biased vectorized prior. We rasterize them with an operator $\mathcal{R}_{r}$ into three\mbox{-}channel rasters $M_{R\mathrm{gt}},M_{R\mathrm{err}}\in\{0,1\}^{H\times W\times 3}$,
\begin{subequations}
\label{eq:map}
\begin{align}
M_{R\mathrm{gt}} &= \mathcal{R}_{r}(M_{\mathrm{gt}}),\\
M_{R\mathrm{err}} &= \mathcal{R}_{r}(M_{\mathrm{err}}).
\end{align}
\end{subequations}
The subscript $R$ indicates the rasterized domain. We partition the raster grid into non\mbox{-}overlapping patches $h$ and assign each patch is composed of grids $u$. Grids label $L(u)$ is that summarizes spatial agreement between $M_{\mathrm{Rgt}}$ and $M_{\mathrm{Rerr}}$,
\begin{equation}
L(u)=
\begin{cases}
\texttt{empty},    & M_{R\mathrm{gt}}(u)=0\ \wedge\ M_{R\mathrm{err}}(u)=0,\\
\texttt{erroneous},& M_{R\mathrm{gt}}(u)\neq M_{R\mathrm{err}}(u),\\
\texttt{reliable}, & \text{otherwise}.
\end{cases}
\label{eq:label}
\end{equation}
Empty grids contain no elements in either map, erroneous patches reflect disagreements, and all remaining occupied patches are treated as reliable.

During training the U\mbox{-}Net receives a single rasterized map as input for each patch. We sample a Bernoulli variable $B(h)\sim\mathrm{Bernoulli}(\alpha)$ and construct a mixed raster $M_{R}$. It means that we mix in the patch level but the label is at the grid level.
\begin{equation}
M_{R}(h)=B(h)\,M_{R\mathrm{err}}(h)+\bigl(1-B(h)\bigr)\,M_{R\mathrm{gt}}(h),
\label{eq:mix}
\end{equation}
which randomly selects either the biased prior or the ground\mbox{-}truth prior in that patch. Labels are always computed from the pair $(M_{R\mathrm{gt}},M_{R\mathrm{err}})$, so supervision is decoupled from input mixing and training remains stable. To address the class imbalance issue, we use focal loss and train with the main model. At inference time PACE only observes the available prior map, for example the biased prior in our experiments. We refer to $M_{\mathrm{conf}}$ as the confidence map and later derive raster masks and scalar confidence fields from it in DPQG and CGPF.

\subsection{Dual\mbox{-}Prior Query Generator (DPQG)}

The confidence estimator identifies unreliable regions in the prior map, but an imperfect prior still contains valuable topological structure, such as lane connectivity and divider layout. The Dual\mbox{-}Prior Query Generator (DPQG) extracts this structure and converts it into stable queries while avoiding regions that PACE marks as unreliable.

\paragraph{Vectorized map embedding.}
The biased vectorized map $M_{\mathrm{err}}$ contains $N$ instances, each represented as a sequence of points. For the $n$\mbox{-}th point $p_n$, we denote its BEV coordinate by $\mathbf{g}_{n}=(x_n,y_n)$, its semantic type by $c_n$, the index of the instance it belongs to by instance $i_n$, and its order in that instance by $s_n$. We map $\mathbf{g}_{n}$ to the nearest raster grid and denote this grid location by $u_n$. Each point $p_n$ is converted into a high\mbox{-}dimensional embedding
\begin{equation}
\mathbf{e}_{n}
= E_{\mathrm{inst}}(i_{n})
+ E_{\mathrm{point}}(s_{n})
+ \mathrm{PE}(u_{n})
+ E_{\mathrm{type}}(c_{n}),
\end{equation}
where $E_{\mathrm{inst}}$ and $E_{\mathrm{point}}$ are learnable instance and order embeddings, $\mathrm{PE}$ is a 2D positional embedding shared with the raster branch as in HRMapNet, and $E_{\mathrm{type}}$ assigns a semantic code to each vector category.

Following PriorDrive \cite{zeng2024driving}, we apply hierarchical attention over these point embeddings. Intra\mbox{-}instance attention operates within each instance to capture spatial geometry, and inter\mbox{-}instance attention operates across instances to model global topology. The output is a tensor $F_{p}^{\mathrm{emb}}\in\mathbb{R}^{N\times M\times C}$, where $M$ is the maximum number of points per instance (with padding when needed) and $C$ is the feature dimension.

\paragraph{Confidence\mbox{-}guided masking.}
Standard encoders treat all points as equally reliable, which allows errors in the prior to propagate. DPQG instead uses the confidence map $M_{\mathrm{conf}}$ predicted by PACE. For each grid $u$, PACE predicts a confidence map $M_{\mathrm{conf}}\in\mathbb{R}^{H\times W\times 3}$ with three channels for reliable, erroneous, and empty. After a channel\mbox{-}wise softmax, we denote these probabilities by $m_{\mathrm{rel}}(u)$, $m_{\mathrm{err}}(u)$, and $m_{\mathrm{empty}}(u)$. We first compute an occupancy score
\begin{equation}
o(u)=1-m_{\mathrm{empty}}(u), \qquad o\in \mathbb{R}^{H\times W},\ 0 \le o \le 1.
\end{equation}
A grid is treated as a valid prior grid only when the occupancy is above a fixed threshold and the reliable probability exceeds the erroneous one, which yields a binary raster mask
\begin{equation}
C_{\mathrm{mask}}^{R}(u)=
\begin{cases}
1, & o(u) > \tau \ \text{and}\ m_{\mathrm{rel}}(u) > m_{\mathrm{err}}(u),\\
0, & \text{otherwise}.
\end{cases}
\end{equation}
Grids that are empty or strongly biased are thus removed from the prior.

We map this raster mask to vector points through their grid locations. Each point coordinate $\mathbf{g}_{n}$ uses the mask value of its grid $u_n$, giving a grid\mbox{-}to\mbox{-}instance mask $C_{\mathrm{mask}}\in\{0,1\}^{N\times M}$. Let $i$ index instances, $m$ index points within an instance, and $c$ index channels. After the attention encoder we apply this mask to the point features by
\begin{equation}
\hat{F}_{p}^{\mathrm{emb}}(i,m)
=
C_{\mathrm{mask}}(i,m)\,F_{p}^{\mathrm{emb}}(i,m).
\end{equation}
So that points in unreliable or empty regions are zeroed out, while points in valid prior regions keep their features. Flattening along the instance and point dimensions yields a topology\mbox{-}aware vector feature matrix $F_{\mathrm{vec}}\in\mathbb{R}^{N_{\mathrm{vec}}\times C}$, where $N_{\mathrm{vec}}=N\times M$, which serves as vectorized prior tokens.

\paragraph{Dual\mbox{-}prior fusion and queries.}
DPQG also uses a rasterized prior branch so that the decoder can access both vectorized and rasterized evidence. Following HRMapNet \cite{zhang2024enhancing}, we embed the rasterized prior into a feature map $F_{\mathrm{Ras}}\in\mathbb{R}^{H\times W\times C}$ using map prior embedding. It has position and label embeddings at each vaild grid $u$, and $l(u)$ is label of grid $u$:
\begin{equation}
F_{\mathrm{Ras}}(u) = \mathrm{PE}(u) + \mathrm{LE}(l(u)).
\end{equation}
Here $\mathrm{PE}(u)$ shares the positional function with the vector branch and keeps both in the same BEV frame, while $\mathrm{LE}(u)$ is a per\mbox{-}class label embedding on raster grids, analogous to $E_{\mathrm{type}}$ for vector categories. The vaild grid is also obtained from $C_{\mathrm{mask}}^{R}$.
So only confident prior regions contribute to the raster feature map.
We reshape $F_{\mathrm{Ras}}$ into a set of raster tokens named map query and concatenate them with the vector tokens named instance query to form a fused prior feature $F_{\mathrm{fuse}}\in\mathbb{R}^{N_{\mathrm{fuse}}\times C}$, where $N_{\mathrm{fuse}}$ is the total number of tokens from both sources.

Finally, we update the learnable base queries $Q_{\mathrm{init}}$ by cross attention over this fused prior feature,
\begin{equation}
Q_{\mathrm{map}} = \operatorname{CrossAttention}\!\big(Q_{\mathrm{init}},\,F_{\mathrm{fuse}},\,F_{\mathrm{fuse}}\big),
\end{equation}
obtaining prior\mbox{-}aware queries $Q_{\mathrm{map}}$ that encode both vector topology and raster layout while being restricted to confident prior regions. These queries are then fed to the map decoder, which can attend to informative prior structures while staying robust to biased parts of the prior map.

\subsection{Confidence\mbox{-}Guided Prior Fusion (CGPF)}

\begin{figure}[t]
  \centering

    \begin{subfigure}{0.46\linewidth}
        \includegraphics[width=\linewidth]{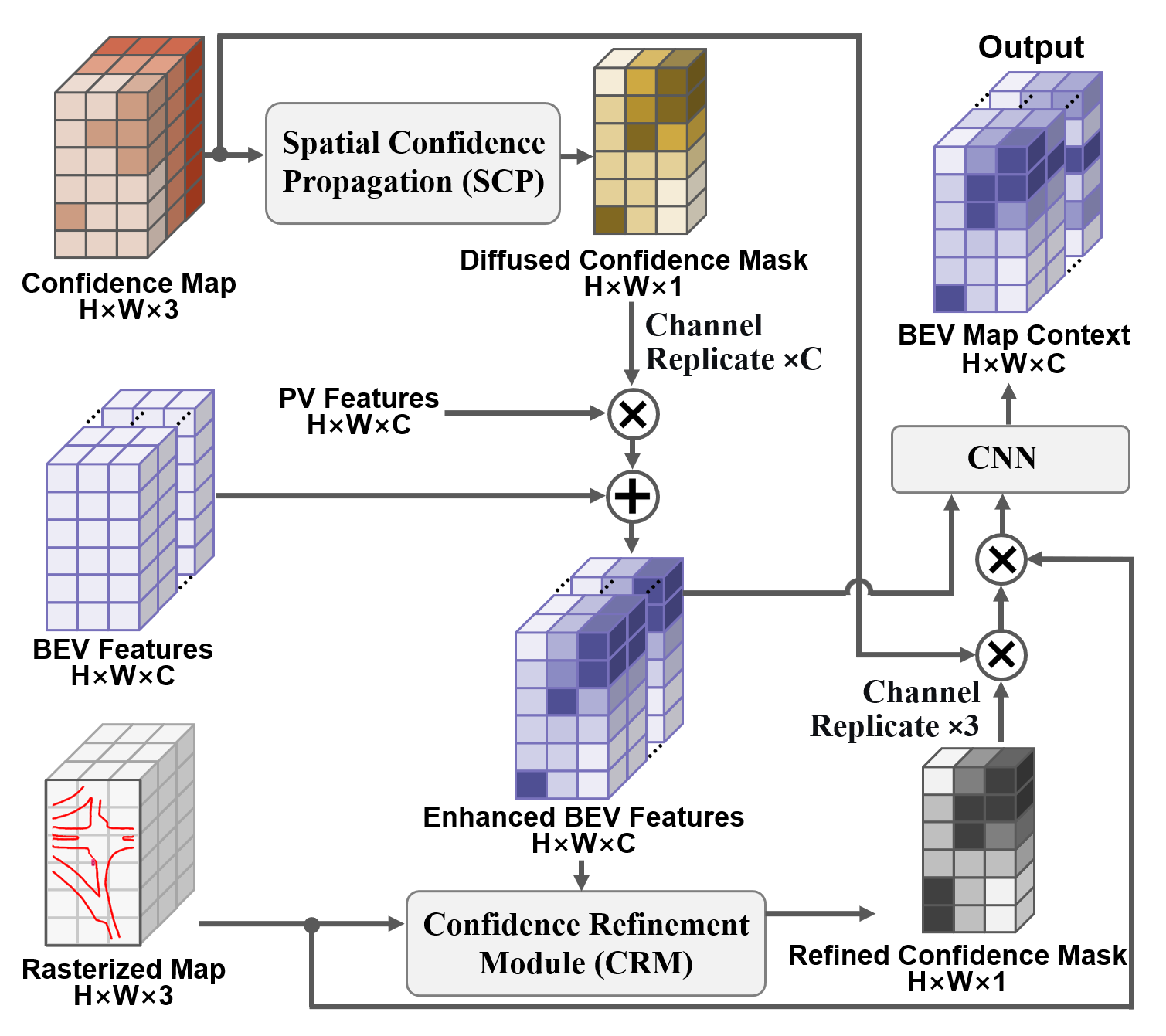}
        \caption{The architecture of CGPF}
        \label{fig:short-a}
      \end{subfigure}
    \hspace{1cm}
    \begin{subfigure}{0.36\linewidth}
        \includegraphics[width=\linewidth]{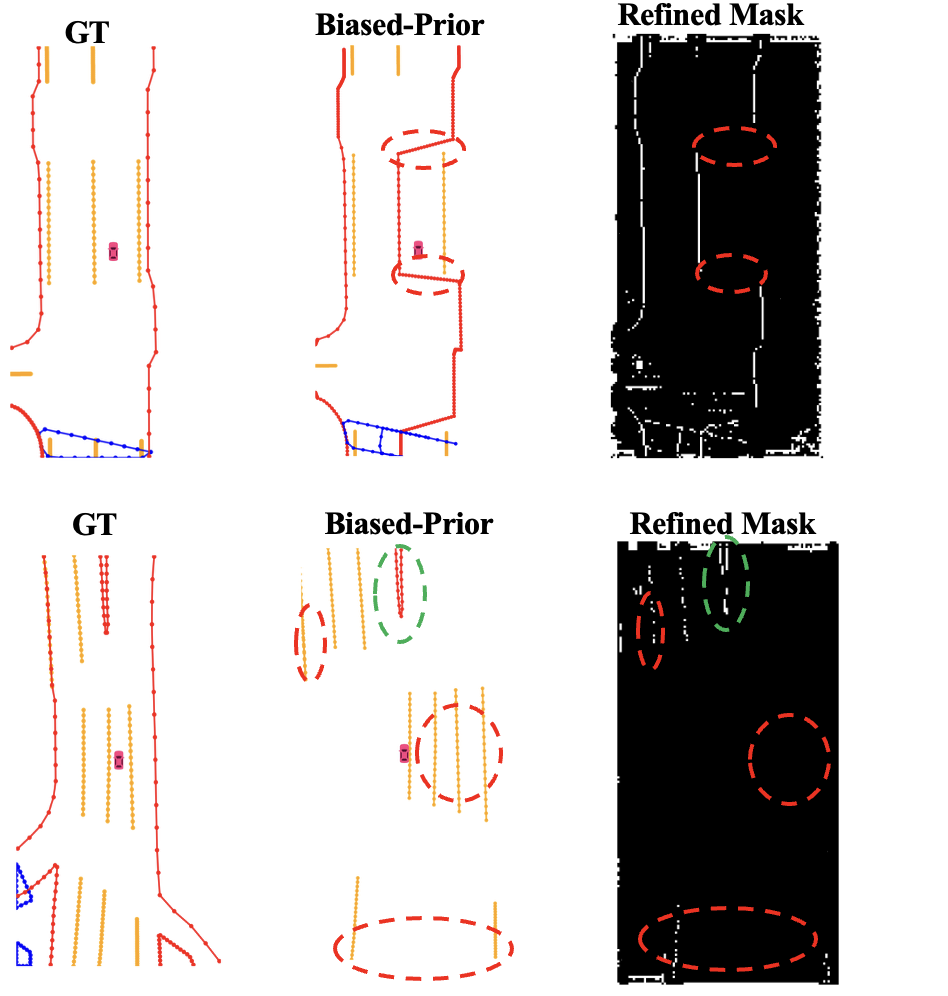}
        \caption{Illustration of the binarized \protect\\
        refined confidence mask.}
        \label{fig:short-b}
    \end{subfigure}
      \caption{(a) Architecture of the Confidence-Guided Prior Fusion (CGPF) module. It contains a Spatial Confidence Propagation (SCP) block that produces a diffused confidence mask and a Confidence Refinement Module (CRM) that refines prior weights and outputs the BEV map context. (b) Illustration of the binarized refined confidence mask. For clarity, it is presented as a binary map where white indicates reliable regions of the prior and black indicates erroneous ones. Compared to the Ground Truth (GT), erroneous regions in the Biased Prior (red dashed circles) are excluded as black in the map, while reliable regions (green dashed circles) are preserved as white.}
      \label{fig:short}
    \label{fig:Innovation3}
\end{figure}

While DPQG focuses on vector queries that capture topology, nearby grids on the BEV grid also provide strong cues. Errors in the prior often appear as small gaps, offsets, or short spatial shifts. The Confidence\mbox{-}Guided Prior Fusion (CGPF) module uses the confidence map to enlarge reliable regions in a controlled way and to decide how prior maps and visual features contribute to the final BEV context as show in \cref{fig:short-a}.

We operate on three BEV aligned inputs. The BEV feature map from the image backbone is $F_{\mathrm{BEV}}\in\mathbb{R}^{H\times W\times C}$. Multi\mbox{-}view perspective features are processed and yields another BEV aligned feature map $F_{\mathrm{PV}}\in\mathbb{R}^{H\times W\times C}$. The biased rasterized prior is $M_{\mathrm{Rerr}}\in\mathbb{R}^{H\times W\times 3}$.
\paragraph{Spatial Confidence Propagation (SCP).}
SCP builds a diffused confidence mask through a dilation\mbox{-}like operation on the BEV grid. We first ignore grids that are likely to be empty and retain only those where $m_{\mathrm{empty}}$ is below a fixed threshold. On these grids, we measure the fraction of reliable mass among all non\mbox{-}empty states,
\begin{equation}
r(u)=
\begin{cases}
\dfrac{m_{\mathrm{rel}}(u)}
{m_{\mathrm{rel}}(u)+m_{\mathrm{err}}(u)+\varepsilon},
& m_{\mathrm{empty}}(u) < \tau_{\mathrm{empty}},\\[6pt]
0, & \text{otherwise},
\end{cases}
\end{equation}
where $\tau_{\mathrm{empty}}$ is a fixed empty threshold and $\varepsilon$ is a small constant for numerical stability. The scalar field
$r \in \mathbb{R}^{H\times W}$, with $0 \le r \le 1$, summarizes the degree of correctness conditioned on the prior being non\mbox{-}empty.

To capture different reliability levels, we split $r$ into $T$ confidence bands. Each band is described by a selector
$S_{\ell}(r)\in \mathbb{R}^{H\times W}$, with
$0 \le S_{\ell}(r) \le 1$, which highlights grids with similar reliability for $\ell=1,\dots,T$. For each band, we apply a diffusion operator $D_{\kappa_{\ell}}$ with kernel parameter $\kappa_{\ell}$ and assign a non\mbox{-}negative weight $w_{\ell}$. The diffused confidence mask is
\begin{equation}
D_m(u)
=
\sum_{\ell=1}^{T}
w_{\ell}\,
\big[D_{\kappa_{\ell}}(S_{\ell}(r))\big](u).
\end{equation}
Here, $D_{\kappa_{\ell}}$ is implemented as convolution with a compact averaging kernel. Higher reliability bands use tighter kernels and larger weights, whereas lower reliability bands use wider kernels to enlarge the spatial search range and smaller weights for conservative propagation. Thus, the lower magnitude of $D_m$ in unreliable regions is a consequence of conservative weighting rather than the purpose of using wider kernels. This soft morphological diffusion enlarges reliable support around the prior and bridges small gaps while retaining a conservative shape.

The diffused mask $D_m$ then controls how multi\mbox{-}view evidence is injected into the BEV representation. We broadcast $D_m$ along the channel dimension and compute
\begin{equation}
F_{\mathrm{enh}}
=
F_{\mathrm{BEV}}
+
D_m\odot F_{\mathrm{PV}},
\qquad
F_{\mathrm{enh}}\in\mathbb{R}^{H\times W\times C},
\end{equation}
where $\odot$ denotes element\mbox{-}wise multiplication. Importantly, $D_m$ modulates only the additional $F_{\mathrm{PV}}$ branch, while $F_{\mathrm{BEV}}$ remains unscaled. A small $D_m(u)$ reduces the additional support from $F_{\mathrm{PV}}(u)$, whereas a large $D_m(u)$ injects more support; in both cases, the online BEV evidence is preserved.

\paragraph{Confidence Refinement Module (CRM).}
SCP propagates spatial reliability, but it does not determine whether the prior content agrees with the visual evidence. The Confidence Refinement Module (CRM) refines the prior weights by evaluating this agreement.

CRM takes the enhanced BEV feature $F_{\mathrm{enh}}$ and the rasterized prior $M_{\mathrm{Rerr}}$ as input. We concatenate them along the channel dimension and apply a shallow convolutional network $\phi$, followed by a sigmoid function, to produce a single\mbox{-}channel weight:
\begin{equation}
G(u) = \sigma\!\big(\phi([\,F_{\mathrm{enh}},\,M_{\mathrm{Rerr}}\,])(u)\big),
\qquad
G \in \mathbb{R}^{H\times W}.
\end{equation}
Unlike the early, grid\mbox{-}wise reliability estimate $m_{\mathrm{rel}}$ used by SCP, $G$ is a late\mbox{-}stage fusion weight that incorporates local spatial context through $\phi$. It is high where the prior geometry is consistent with the enhanced BEV feature and low where they disagree. We combine it with $m_{\mathrm{rel}}(u)$ to form the refined confidence mask
\begin{equation}
\beta(u) = m_{\mathrm{rel}}(u)\,G(u),
\qquad
\beta\in\mathbb{R}^{H\times W}.
\end{equation}
Grids that are reliable but inconsistent with the images are downweighted, whereas grids that are both reliable and consistent retain large weights.

Finally, the refined mask shown in \cref{fig:short-b} reweights only the rasterized prior branch. We broadcast $\beta$ along the channel dimension, concatenate the weighted prior $\beta\odot M_{\mathrm{Rerr}}$ with the unscaled $F_{\mathrm{enh}}$, and fuse them using a shallow CNN to obtain the BEV map context
$F_{\mathrm{ctx}}\in\mathbb{R}^{H\times W\times C}$. This map context serves as the key and value for the map decoder. Consequently, low\mbox{-}confidence or conflicting regions reduce only the contribution of the prior map, while the online visual evidence in $F_{\mathrm{enh}}$ remains preserved. In this way, CGPF provides a confidence\mbox{-}aware BEV representation that can be readily integrated into standard decoders.

\section{Experiments}
\subsection{Datasets}
\paragraph{Base datasets.}
We evaluate on the nuScenes \cite{caesar2020nuscenes} benchmarks for online vectorized map construction. 
nuScenes contains $1\,000$ urban driving scenes (about $20$\,s each) recorded at $2$\,Hz with six surrounding RGB cameras that cover a full $360^\circ$ field of view. we use the official split of scenes $700/150/150$ for training, validation, and testing. 

\paragraph{Biased maps.}
To train models that are robust to biased priors, we consider both real and synthetic biased maps. 
Although Argoverse2 \cite{wilson2023argoverse} includes a small set of biased maps, their scale is insufficient for our study, so we mainly rely on synthetic priors. 
Among synthetic options, MapEX \cite{Sun_2025_WACV} and Syn\mbox{-}D\mbox{-}Maps \cite{11097394} target nuScenes. 
MapEX defines several error categories (e.g., missing content, global translation, local warping) but does not model explicit topological changes and is not publicly available. 
Syn\mbox{-}D\mbox{-}Maps, in contrast, is an open\mbox{-}source pipeline that supports topology\mbox{-}altering operations such as lane widening/narrowing and lane addition/reduction. 
Following its procedure on nuScenes, we generate four topology-altering biased types (lane widening, lane narrowing, lane addition, lane reduction) along with rotation biased type, yielding one biased prior per types for every frame.

\subsection{Implementation Details}\label{Exp_imp}
\paragraph{Metric.}
For the nuScenes dataset, we use the official split of scenes. 
Following previous work, we evaluate on validation set and report performance on three categories. We report class-wise AP and mAP following the official nuScenes protocol, which matches polylines using Chamfer-distance thresholds of $\{0.5,1.0,1.5\}\,\mathrm{m}$. 
The prediction map range is $[-15,15]\,\mathrm{m}$ along the $X$-axis and $[-30,30]\,\mathrm{m}$ along the $Y$-axis.

\paragraph{Generate Biased Maps.}
We adopt Syn-D-Maps\cite{11097394} as our corruption framework to introduce five types of map bias. These consist of four topology-altering biases (lane addition, lane reduction, lane widening, and lane narrowing) and a rotation bias.
When they modify maps, all related instances (e.g. road boundaries) are modified accordingly to preserve topological consistency. 
In accordance with the Syn-D-Maps setup, we applied a corruption ratio of 0.6 for four topology-altering biased types, defined as the fraction of total lanes subject to modification. For rotation biased type, it controls bias by varying the rotation angles. We also follow the setup from Syn-D-Maps.
During training, each sample is paired with one biased types drawn uniformly from four topology-altering biased types. The rotation biased type is used in ablation study.


\paragraph{Model.}We adopt a ResNet\mbox{-}50 backbone with a total batch size of $32$. The base learning rate is set to $6\times10^{-4}$ and we use a learning rate of $1.2\times10^{-3}$ for the U\mbox{-}Net branch. We follow the loss function from MapTRv2 \cite{liao2025maptrv2}, which combines classification, point-to-point, and edge direction losses, along with auxiliary one-to-many loss \cite{Jia_2023_CVPR}, dense prediction, and depth losses. We add Focal Loss \cite{Lin_2017_ICCV} to train the U-Net.

In training, we set $\alpha = 0.5 $, $N=35$ and $M=20$. All retrained models (ours and $\dagger$ baselines) are trained for 24 epochs on eight NVIDIA A100 GPUs and the remaining training settings follow HRMapNet. For the three baseline methods that rely on priors or temporal history, we keep their original training protocols and only replace the original priors or historical maps with the biased priors. All models share the same data mix strategy described in the Methods section, and during evaluation they access only the biased maps.
\begin{table}[t]
  \centering
  \caption{Comparison with state\mbox{-}of\mbox{-}the\mbox{-}art methods on the nuScenes dataset.
  The first block lists methods that do not rely on priors or history for map
  construction, while the second block lists methods that leverage various priors. The third block lists models retrained and evaluated only with biased priors under our mixed-bias setting. 
  $^\ast$ Err. denotes that, during evaluation, only biased prior maps are used as inputs. $\dagger$ The results are trained with our mixed data from scratch by replacing history or prior maps.}
  \label{tab:comparison}
  \renewcommand{\arraystretch}{0.75} 
  \begin{threeparttable}
  \footnotesize
  \begin{tabular}{lcccccccc}
    \toprule
    Methods & Backbone & Epoch & Prior & Err.$^\ast$ &
    $AP_{\text{ped}}$ & $AP_{\text{div}}$ & $AP_{\text{boun}}$ & mAP \\
    \midrule
    StreamMapNet \cite{yuan2024streammapnet}~\textcolor{gray}{WACV24} & R50 & 30  & --          & --          & 56.9 & 55.9 & 61.4 & 58.1 \\
    MGMap \cite{liu2024mgmap}~\textcolor{gray}{CVPR24}          & R50 & 24  & --          & --          & 61.8 & 65.0 & 67.5 & 64.8 \\
    HIMap \cite{zhou2024himap}~\textcolor{gray}{CVPR24}         & R50 & 30  & --          & --          & 62.6 & 68.4 & 69.1 & 66.7 \\
    MapQR \cite{liu2024leveraging}~\textcolor{gray}{ECCV24}     & R50 & 24  & --          & --          & 63.4 & 68.0 & 67.7 & 66.4 \\
    Mask2Map \cite{choi2024mask2map}~\textcolor{gray}{ECCV24}   & R50 & 24  & --          & --          & 70.6 & 71.3 & 72.9 & 71.6 \\
    MapTRv2 \cite{liao2025maptrv2}~\textcolor{gray}{IJCV25}     & R50 & 24  & --          & --          & 59.8 & 62.4 & 62.4 & 61.5 \\
    DAMap \cite{dong2025damap}~\textcolor{gray}{ICCV25}     & R50 & 24  & --          & --          & 70.7 & 72.3 & 74.9 & 72.6 \\
    \midrule
    SQD-MapNet \cite{wang2024stream}~\textcolor{gray}{ECCV24}   & R50 & 24  & \checkmark  & --          & 63.0 & 62.5 & 63.3 & 63.9 \\
    P-MapNet \cite{jiang2024p}~\textcolor{gray}{RAL24}          & EN-B0 & 10  & \checkmark & --   & 41.3 & 54.2 & 63.7 & 53.1 \\
    MapTracker \cite{chen2024maptracker}~\textcolor{gray}{ECCV24} & R50 & 72 & \checkmark  & --          & 80.0 & 74.1 & 74.1 & 76.1 \\
    PriorMapNet \cite{wang2024priormapnet}~\textcolor{gray}{arXiv24} & R50 & 24 & \checkmark & --         & 64.0  & 69.0 & 68.2 & 67.1 \\
    PriorDrive \cite{zeng2024driving}~\textcolor{gray}{arXiv24} & R50 & 24  & \checkmark  & --          &63.6 &67.3 &66.2 &65.7 \\
    HRMapNet \cite{zhang2024enhancing}~\textcolor{gray}{ECCV24} & R50 & 24  & \checkmark  & --          & 65.8 & 67.4 & 68.5 & 67.2 \\
    PrevPredMap \cite{peng2025prevpredmap}~\textcolor{gray}{WACV25} & R50 & 24 & \checkmark & --         & 68.7 & 66.0 & 68.3 & 67.6 \\
    UniPrevPred \cite{peng2025uni}~\textcolor{gray}{arXiv25}    & R50 & 24  & \checkmark  & --          & 76.2 & 72.3 & 73.6 & 74.0 \\
    PriorFusion \cite{tang2025priorfusion}~\textcolor{gray}{arXiv25} & R50 & 110 & \checkmark & --        & 68.3 & 72.0 & 70.9 & 70.4 \\
    \midrule
    SQD-MapNet$^{\dagger}$                                      & R50 & 24  & --  & \checkmark  & 55.4 & 53.3  & 57.9 & 55.5 \\
    PrevPredMap$^{\dagger}$                                     & R50 & 24  & --  & \checkmark  & 63.5 & 62.9  & 66.4 & 64.3 \\
    HRMapNet$^{\dagger}$                                        & R50 & 24  &  --          & \checkmark  & \textbf{78.5} &  82.0& 78.5 & 79.7 \\
    \rowcolor{gray!20}
    PriorMaskMap (ours)                                         & R50 & 24  & --          & \checkmark  &78.1  &  \textbf{85.1} & \textbf{80.9} & \textbf{81.4} \\
    HRMapNet$^{\dagger}$                                        & R50 & 110  &  --          & \checkmark  & \textbf{81.8} &  85.0& 83.3 & 83.4 \\
    \rowcolor{gray!20}
    PriorMaskMap (ours)                                         & R50 & 110  & --          & \checkmark  &81.7  &  \textbf{88.9} & \textbf{84.5} & \textbf{85.0} \\
    \bottomrule
  \end{tabular}
  \end{threeparttable}
\end{table}
\subsection{Main Results}
In ~\cref{tab:comparison}, we summarizes the comparison on nuScenes. 
The first block contains methods that do not use priors or temporal history and rely only on current sensor observations. 
These models already achieve strong performance with mAP up to 72.6. 
In contrast, the last block evaluates models that receive only biased priors at test time under our mixed bias setting. 
PriorMaskMap reaches 81.4 mAP in this regime and still outperforms all no-prior baselines, which shows that the network can extract useful structural cues even when the prior is systematically corrupted.

\begin{figure}[t]
  \centering
   \includegraphics[width=0.85\linewidth]{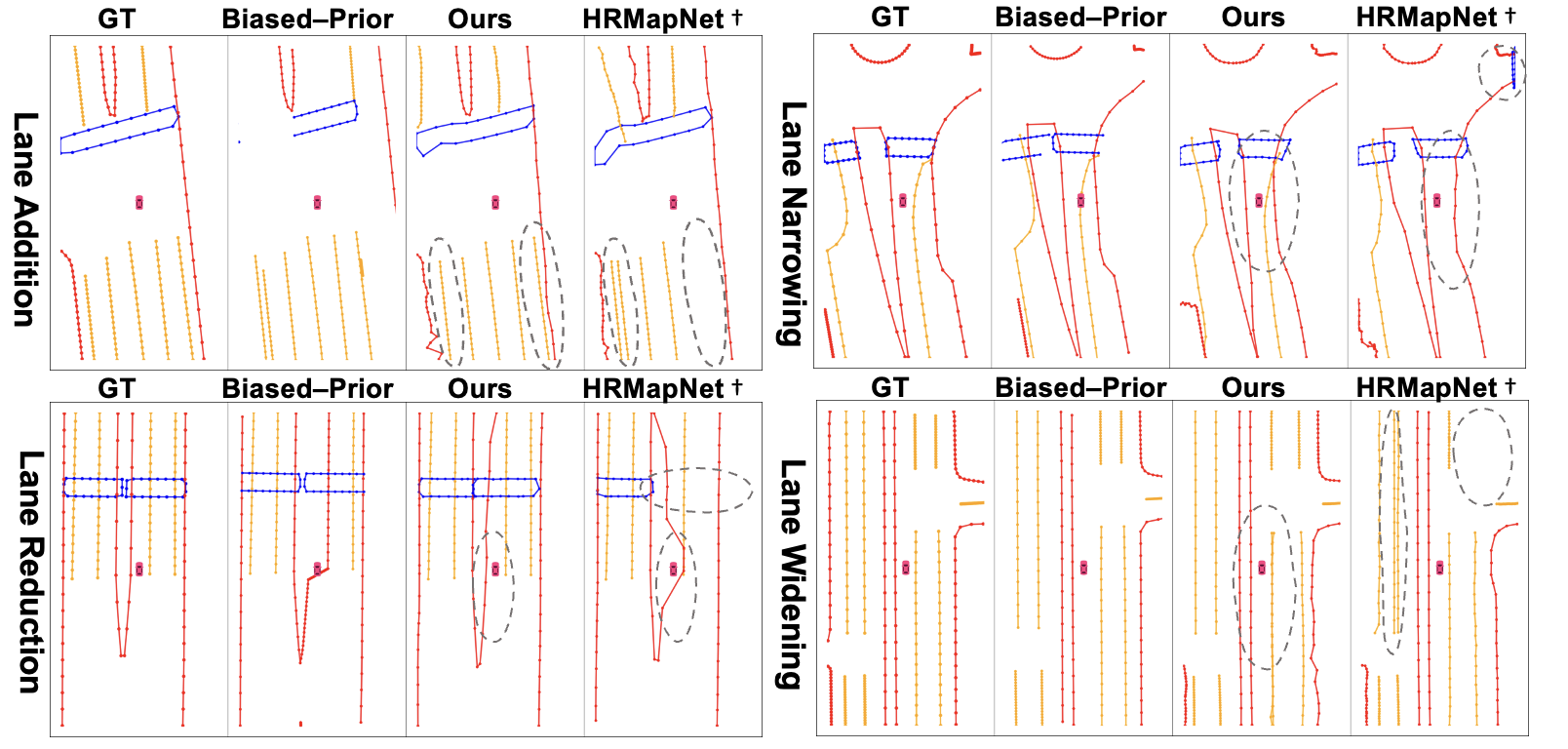}
   \caption{Visualization comparing our proposed model against the HRMapNet$^\dagger$.}
   \label{fig:vis}
\end{figure}

The second block reports approaches that are designed for clean priors or propagated maps and are evaluated in their original configurations with mAP around 70 to 76. 
To place prior-based models in the same biased regime we retrain SQD\mbox{-}MapNet$^{\dagger}$, PrevPredMap$^{\dagger}$ and HRMapNet$^{\dagger}$ with the mixed data which gives the results in the last block. 
Among these baselines HRMapNet$^{\dagger}$ is the strongest with 79.7 mAP while PriorMaskMap further improves mAP to 81.4 without access to any clean prior or temporal history.
Additionally, we compare our model against HRMapNet$^{\dagger}$  over a longer training duration. Even at epoch 110, our approach continues to sustain an advantage of 1.6 mAP.
This suggests that explicitly modeling prior reliability recovers most of the benefit of a prior and even surpasses prior-based baselines once the maps are biased, some examples are shown in \cref{fig:vis}. 

\paragraph{Impact of topology-altering biased types.}
We further analyze the impact of each topology-altering bias in ~\cref{tab:types-results}. Compared with HRMapNet$^{\dagger}$, PriorMaskMap consistently achieves higher mAP across all four biased types, with modest gains for lane addition and reduction and noticeably larger improvements for lane widening and narrowing. Most of the benefit comes from higher AP on lane dividers and road boundaries, while pedestrian-crossing AP stays within about $\pm 1$ point relative to HRMapNet$^{\dagger}$. Meanwhile, SQD-MapNet$^{\dagger}$ and PrevPredMap$^{\dagger}$ show almost unchanged performance across different biased types, which we attribute to their conservative use of prior information: when the bias becomes large, the models tend to largely ignore the prior. Overall, these results indicate that the proposed confidence-guided diffusion and topology-aware aggregation can effectively recover missing or distorted topology under diverse structure-changing corruptions instead of overfitting to a specific biased pattern.
\begin{table}[t]
  \centering
  \caption{Performance comparison on the nuScenes dataset with mixed bias prior. $^\dagger$ as in ~\cref{tab:comparison} and the first two results are not typos but follow from their respective processing mechanisms. Bold numbers indicate the best mAP for each biased type.}
  \label{tab:types-results}
  \setlength{\tabcolsep}{2.5pt}
  \renewcommand{\arraystretch}{0.75}
  \small
  \begin{threeparttable}
  \begin{tabular}{@{\extracolsep{\fill}}ll c rrr r}
  \toprule
  Method & Biased Type & Epoch &
  $AP_{\text{ped}}$ & $AP_{\text{div}}$ & $AP_{\text{boun}}$ & mAP \\
  \midrule
    \multirow{4}{*}{SQD-MapNet$^{\dagger}$} &
       Addition  & \multirow{4}{*}{24} & 55.4 & 53.3 & 57.9 & 55.5 \\
    &  Reduction &                    & 55.4 & 53.3 & 57.9 & 55.5 \\
    &  Widening  &                    & 55.4 & 53.4 & 57.9 & 55.6 \\
    &  Narrowing &                    & 55.4 & 53.3 & 57.9 & 55.5 \\
    \midrule
    \multirow{4}{*}{PrevPredMap$^{\dagger}$} &
       Addition  & \multirow{4}{*}{24} & 63.1 & 60.6 & 65.9 & 63.2 \\
    &  Reduction &                    & 63.1 & 60.6 & 65.9 & 63.2 \\
    &  Widening  &                    & 63.1 & 60.6 & 65.9 & 63.2 \\
    &  Narrowing &                    & 63.1 & 60.6 & 65.9 & 63.2 \\
    \midrule
    \multirow{4}{*}{HRMapNet$^{\dagger}$} &
       Addition  & \multirow{4}{*}{24} & 68.2 & 77.6 & 68.9 & 71.6 \\
    &  Reduction &                    & 85.4 & 93.8 & 87.6 & 88.9 \\
    &  Widening  &                    & 76.7 & 75.1 & 74.2 & 75.3 \\
    &  Narrowing &                    & 77.7 & 75.6 & 75.2 & 76.2 \\
    \midrule
    \multirow{4}{*}{Ours} &
       Addition  & \multirow{4}{*}{24} & 67.3 & 79.9 & 69.5 & \textbf{72.2} \\
    &  Reduction &                    & 86.3 & 94.6 & 87.3 & \textbf{89.4} \\
    &  Widening  &                    & 76.3 & 81.3 & 79.5 & \textbf{79.0} \\
    &  Narrowing &                    & 76.7 & 80.9 & 78.1 & \textbf{78.9} \\
    \bottomrule
  \end{tabular}
  \end{threeparttable}
\end{table}
\paragraph{Result of spatially non-overlapping splits}
Recent work suggests that the official nuScenes split is suboptimal for the online vectorized map construction task, as overlapping geographical regions inherently lead to model overfitting. Therefore, to more comprehensively evaluate our model's performance in real-world scenarios, we conduct experiments on two widely adopted alternative splits proposed by StreamMapNet \cite{yuan2024streammapnet} and the Localization Is All You Evaluate \cite{lilja2024localization} (Geo-split) in \cref{tab:nonoverlap}. While a performance decline is observed across the board, our model continues to outperform HRMapNet$^\dagger$. Specifically, we achieve a 3.4 mAP advantage under the Geo-split, which further expands to a 4.9 mAP margin on the StreamMapNet split. Overall, these results demonstrate our model's robust generalization capabilities and consistent superiority in unseen environments.

\begin{table}[t]
  \centering
  \caption{Quantitative comparison with the baseline on StreamMapNet and Geo-split datasets. $^\dagger$ as in ~\cref{tab:comparison}. Best results are highlighted in \textbf{bold}. Both models are trained for 24 epochs.}
  \label{tab:nonoverlap}
  \begin{tabular}{lcccccccc}
    \toprule
    \multirow{2}{*}{Method} & \multicolumn{4}{c}{StreamMapNet} & \multicolumn{4}{c}{Geo-split} \\
    \cmidrule(lr){2-5} \cmidrule(lr){6-9}
    & $AP_{\text{ped}}$ & $AP_{\text{div}}$ & $AP_{\text{boun}}$ & mAP & $AP_{\text{ped}}$ & $AP_{\text{div}}$ & $AP_{\text{boun}}$ & mAP \\
    \midrule
    HRMapNet$^{\dagger}$ & 38.0 & 70.0 & 71.3 & 59.7 & 51.9 & 65.7 & 64.7 & 60.8 \\
    Ours     & 40.2 & 78.7 & 74.8 & \textbf{64.6}(+4.9) & 49.9 & 75.5 & 67.3 & \textbf{64.2}(+3.4) \\
    \bottomrule
  \end{tabular}
\end{table}

\subsection{Ablation study} 
We analyzes the contribution of each component in ~\cref{tab:ablation}. Compared to topology-altering biased types, the rotation biased type shifts all map elements from their original positions, thereby better highlighting the performance gaps between different component. The result shows that removing vectorized prior in DPQG while keeping SCP and CRM leads to the largest degradation, reducing mAP from $76.7$ to $74.8$, which shows that topology-aware prior queries are crucial for exploiting biased maps. However, retaining only SCP and CRM still yielded a 3.4 mAP improvement. This clearly demonstrates the effectiveness of both modules.


Starting from the DPQG-only configuration, enabling either CRM or SCP alone brings consistent improvements of around $0.6$\mbox{-}$0.7$ mAP and slightly boosts both divider and boundary AP. 
Combining all three modules yields the best performance with $76.7$ mAP, indicating that SCP and CRM provide complementary spatial diffusion and local refinement on top of DPQG. 
Overall, DPQG supplies informative prior tokens, while SCP and CRM further regulate where and how these priors influence the BEV representation.

\begin{table}[t]
  \centering
  \caption{Ablation study of DPQG, SCP, and CRM.}
  \label{tab:ablation}
  \setlength{\tabcolsep}{3.8pt}
  \renewcommand{\arraystretch}{0.95}
  \small
  \begin{tabular}{ccccccc}
    \toprule
    Vectorized Prior  & SCP & CRM & $AP_{\text{ped}}$ & $AP_{\text{div}}$ & $AP_{\text{boun}}$ & mAP \\
    \midrule
    \ding{55}  & \ding{55}          & \ding{55}         & 74.5 & 69.3 & 70.3 & 71.4 \\
    \rowcolor{gray!20} \ding{51}  & \ding{55}          & \ding{55}         & 79.1 & 74.4 & 73.9 & 75.8 (+4.4) \\
    \ding{51}  & \ding{55}          & \ding{51}         & 80.2 & 73.5 & 75.9 & 76.5 \\
    \ding{51}  & \ding{51}          & \ding{55}         & 79.7 & 74.9 & 74.5 & 76.4 \\
    \rowcolor{gray!20} \ding{55}  & \ding{51}          & \ding{51}         & 80.5 & 72.6 & 71.6 & 74.8  (+3.4) \\
    \rowcolor{gray!20} \ding{51}  & \ding{51}          & \ding{51}         & 79.6 & 75.3 & 75.1 & \textbf{76.7} (+5.3) \\
    \bottomrule
  \end{tabular}
\end{table}

\section{Conclusion}
This work presents PriorMaskMap, a robust framework for online vectorized map construction that enhances mapping precision by selectively leveraging biased prior maps. Our approach is built upon a core Prior-Aware Confidence Estimator (PACE) that identifies trustworthy regions in prior maps and produces a grid-wise confidence map. This output directs the subsequent generation of map element queries and feature fusion. Specifically, the Dual-Prior Query Generator (DPQG) creates query embeddings through separate rasterized and vectorized prior branches, while the Confidence-Guided Prior Fusion (CGPF) spatially propagates confidence scores to regulate BEV-prior fusion.

To support evaluation, we extend the nuScenes dataset with synthesized biased prior maps that simulate diverse and realistic structural deviations. The experiments demonstrate that PriorMaskMap achieves superior accuracy while maintaining remarkable robustness under significantly inaccurate priors. Our future work will explore temporal fusion of sequential predictions and updating offline priors with online mapping results to further improve performance. 




%
%
\bibliographystyle{splncs04}
\bibliography{main}

\end{document}